\documentclass{article}
\usepackage{ijcai23}

\usepackage{times}
\usepackage{soul}
\usepackage{url}
\usepackage[hidelinks]{hyperref}
\usepackage[utf8]{inputenc}
\usepackage[small]{caption}
\usepackage{graphicx}
\usepackage{amsmath}
\usepackage{amsthm}
\usepackage{booktabs}
\usepackage{algorithm}
\usepackage{algorithmic}
\usepackage[switch]{lineno}

\usepackage{amsmath}
\usepackage{amssymb}
\usepackage{mathtools}
\usepackage{amsthm}
\usepackage{microtype}
\usepackage{graphicx}
\usepackage{subfigure}
\usepackage{booktabs} 
\usepackage{times}
\usepackage{latexsym}
\usepackage{amsmath}
\usepackage{amssymb}
\usepackage{mathrsfs}
\usepackage{algorithm}
\usepackage{algorithmic}
\usepackage{diagbox}
\usepackage{color}
\usepackage{soul}
\usepackage{multirow} 
\usepackage{stfloats}
\usepackage{siunitx}
\usepackage{url}
\usepackage{hyperref}
\usepackage{etoolbox}

\title{Integrating Local Real Data with Global Gradient Prototypes for Classifier Re-Balancing in Federated Long-Tailed Learning}

\author{
Wenkai Yang$^1$
\and
Deli Chen$^2$\and
Hao Zhou$^{2}$\and
Fandong Meng$^{2}$\and
Jie Zhou$^{2}$
\And
Xu Sun$^{3}$
\affiliations
$^1$Center for Data Science, Peking University\\
$^2$Pattern Recognition Center, WeChat AI, Tencent Inc., China\\
$^3$MOE Key Lab of Computational Linguistics, School of Computer Science, Peking University
\emails
wkyang@stu.pku.edu.cn,
xusun@pku.edu.cn,\\
\{delichen, tuxzhou, fandongmeng, withtomzhou\}@tencent.com
}

\begin{document}

\maketitle

\begin{abstract}
    Federated Learning (FL) has become a popular distributed learning paradigm that involves multiple clients training a global model collaboratively in a data privacy-preserving manner. However, the data samples usually follow a long-tailed distribution in the real world, and FL on the decentralized and long-tailed data yields a poorly-behaved global model severely biased to the head classes with the majority of the training samples. 
    To alleviate this issue, decoupled training has recently been introduced to FL, considering it has achieved promising results in centralized long-tailed learning by re-balancing the biased classifier after the instance-balanced training. 
    However, the current study restricts the capacity of decoupled training in federated long-tailed learning with a sub-optimal classifier re-trained on a set of pseudo features, due to the unavailability of a global balanced dataset in FL. 
    In this work, in order to re-balance the classifier more effectively, 
    we integrate the local real data with the global gradient prototypes to form the local balanced datasets, and thus re-balance the classifier during the local training. 
    Furthermore, we introduce an extra classifier in the training phase to help model the global data distribution, which addresses the problem of contradictory optimization goals caused by performing classifier re-balancing locally. 
    Extensive experiments show that our method consistently outperforms the existing state-of-the-art methods in various settings. 
\end{abstract}

\section{Introduction}
\label{sec: intro}
Federated Learning (FL)~\cite{fedavg} is proposed as an effective distributed learning framework to enable local clients to collaboratively train a global model without exposing their local private data to each other. 
In the real world, there are two data distribution phenomena that introduce great challenges to the good convergence of FL algorithms. One is that the data samples are not identically and independently distributed (non-i.i.d.) across different clients. Furthermore, the other important phenomenon is that the global data distribution (i.e., the data distribution of the training samples merged from all clients' local data) usually shows a long-tailed/class-imbalanced pattern~\cite{long-tailed_learning}, where head classes occupy a much larger proportion of the training samples than tail classes. Directly applying FL on such long-tailed data will produce a global model with poor generalization ability that is severely biased to the head classes~\cite{ratio}. However, it is challenging to deal with FL on the non-i.i.d.\ and long-tailed data due to two aspects. \textbf{First}, affected by the non-i.i.d.\ data partitions, the local data distributions (i.e., local imbalance) show inconsistent long-tailed patterns with that of the global data distribution (i.e., global imbalance)~\cite{ratio}. Thus, tackling the local imbalance problem only (e.g., Fed-Focal Loss~\cite{fed-focal}) will not help to address the global imbalance problem in FL. \textbf{Second}, considering the data privacy concern, it is infeasible to obtain the imbalance pattern of the global data distribution from the local data information. This further limits the application of the global class re-weighting strategy~\cite{effective_number}.

To deal with the above problems, some existing studies manage to estimate the global imbalance pattern by utilizing either the uploaded gradients w.r.t.\ the classifier~\cite{ratio} or the values of local training losses~\cite{climb}. They then apply the class-level~\cite{ratio} or client-level~\cite{climb} re-weighting practice to focus more on the gradients contributed by tail classes or poorly-learned clients. However, previous studies~\cite{decoupling,bbn} have shown that the re-weighting practice will do harm to the representation learning phase. Therefore, the improvement brought by this kind of method is limited. 

Recently, some centralized long-tailed learning studies \cite{decoupling,bbn} manage to decouple the model learning on long-tailed data into the representation learning phase and the classifier learning phase, and find that the instance-balanced training (i.e., uniform sampling on the entire training set to make the contribution of each sample the same) leads to the well-learned representations but a biased classifier. Therefore, centralized decoupled training aims re-train the classifier on a small balanced dataset after the instance-balanced training, and has achieved very promising results. 
However, decoupled training is difficult to be implemented in FL due to the lack of a public balanced dataset. Then, CReFF~\cite{creff} proposes to re-train the classifier on a set of pseudo features created on the server. 
Nevertheless, the improvement brought by CReFF is restricted by the high similarity of the pseudo features per class, and the fundamental problem -- lack of real balanced data still exists.

To better solve the lack of the real balanced data issue in the application of decoupled training in FL, we propose a different yet more effective classifier re-balancing algorithm, and achieve state-of-the-art results in federated long-tailed learning. That is, 
we choose to take full advantage of the abundant real data stored in the local clients, and allow the clients to re-balance the classifier during local training. Specifically, we make each client re-balance the classifier on a local balanced dataset that is mixed with the local real data and the global gradient prototypes of the classifier sent by the server, while the latter is supposed to address the issue of missing classes in the local datasets. 
Additionally, we add an extra classifier in the local training phase to jointly model the global data distribution. This practice helps to overcome the optimization difficulty on the global representation learning brought by the practice of local classifier re-balancing. 
Compared with CReFF, we allow the clients to collaboratively train a balanced classifier with their sufficient real data during local training, which needs no extra requirements on the server and produces an optimal classifier with better generalization ability. We conduct extensive experiments on the three popular long-tailed image classification tasks, and the results show that our method can significantly outperform all existing federated long-tailed learning methods in various settings.

\section{Related Work}
\label{sec: related work}
\subsection{Federated Learning}
Federated Averaging (FedAvg)~\cite{fedavg} is the most widely-used FL algorithm, but it has been shown that the performance of FedAvg drops greatly when the data is non-i.i.d.~\cite{scaffold}. Therefore, plenty of existing FL studies target on dealing with the non-i.i.d.\ data partitions in FL. For example, FedProx~\cite{fedprox} and FedDyn~\cite{feddyn} manage to make the local models converge to the same global optimum by adding the regularization terms in the local training objectives, SCAFFOLD~\cite{scaffold} chooses to correct the local gradient in each step with the gradients from other clients to reduce the gradient variance across different clients. FedAvgM~\cite{fedavgm} and FedOPT~\cite{fedopt} adopt the server momentum and the adaptive server optimizer in the server aggregation phase. 

\subsection{Long-Tailed/Imbalanced Learning}
In the real world, the data points usually show a long-tailed distribution pattern. 
Therefore, learning good models on the long-tailed/class-imbalanced data has been widely studied~\cite{long-tailed_learning} in the traditional centralized learning, and attracts more and more attention in the FL setting.

\paragraph{Centralized Long-Tailed Learning}
The methods to tackle the class imbalance problem in the centralized long-tailed learning can be mainly divided into three categories: (1) \textbf{Class-level re-balancing methods} that includes over-sampling training samples from tail classes~\cite{over-sampling}, under-sampling data points from head classes~\cite{under-sampling}, or re-weighting the loss values or the gradients of different training samples based on the label frequencies~\cite{effective_number,ldam} or the predicted probabilities of the model~\cite{focal-loss}. (2) \textbf{Augmentation-based methods} aim to create more data samples for tail classes either from the perspective of the feature space~\cite{ofa,fasa} or the sample space~\cite{remix}. (3) \textbf{Classifier re-balancing mechanisms} are based on the finding that the uniform sampling on the whole dataset during training benefits the representation learning but leads to a biased classifier, so they design specific algorithms to adjust the classifier during or after the representation learning phase~\cite{bbn,decoupling}.

\paragraph{Federated Long-Tailed Learning}
Recently, a few studies begin to focus on the class imbalance problem in FL, as FL becomes a more practical and popular learning paradigm and the long-tailed data distribution is unavoidable in the real world. Fed-Focal Loss~\cite{fed-focal} directly applies Focal Loss~\cite{focal-loss} in the clients' local training, but it neglects the fact that the local imbalance pattern is inconsistent with the global imbalance pattern. Ratio Loss~\cite{ratio} utilizes an auxiliary dataset on the server (which is usually impractical in real cases) to estimate the global data distribution, and send the estimated information to clients to perform class-level re-weighting during local training. CLIMB~\cite{climb} is proposed as a client-level re-weighting method to give more aggregation weights to the clients with larger local training losses. However, both Ratio Loss and CLIMB bring negative effects to the representation learning caused by the re-weighting practice. FEDIC~\cite{fedic} also needs the impractical assumption to own an auxiliary balanced dataset for fine-tuning the global model on the server, and uses the fine-tuned model along with the local models as teachers to perform knowledge distillation on the original global model. Most recently, CReFF~\cite{creff} adopts the decoupled training idea to re-train the classifier on the server by creating a number of federated features for each class, and achieves previously state-of-the-art performance. However, the low quality and the limited number of federated features restrict its potential.

\section{Methodology}
\subsection{Problem Definition}
\label{subsec: problem definition}
In the FL framework, each client $k$ ($k=1,\cdots,N$) has its own local dataset $\mathcal{D}_{k}$, and all clients form a federation to jointly train a good global model under the constraint that the local data is always kept in the local devices. Then, the optimization goal of FL can be formulated as
\begin{equation}
\resizebox{.89\hsize}{!}{$
\begin{aligned}
    \boldsymbol{\theta}^{*} = \mathop{\arg\min}\limits_{\boldsymbol{\theta}} L(\boldsymbol{\theta}) =\mathop{\arg\min}\limits_{\boldsymbol{\theta}} \sum_{k=1}^{N} \frac{|\mathcal{D}_{k}|}{\sum_{i=1}^{N}|\mathcal{D}_{i}|}L(\boldsymbol{\theta};\mathcal{D}_{k}),
     \end{aligned}
     $}
\end{equation}
where $|\mathcal{D}_{k}|$ represents the total number of training samples in $\mathcal{D}_{k}$, and $L(\cdot;\mathcal{D}_{k})$ is the local training objective in client $k$. 

Federated Averaging (FedAvg)~\cite{fedavg} is the most popular FL framework to solve the above optimization problem. Specifically, at the beginning of each communication round $t$, the server sends the updated global model $\boldsymbol{\theta}^{t-1}$ from the last round to all available/sampled clients $k \in \mathcal{C}^{t}$ in the current round, and each client $k$ takes $\boldsymbol{\theta}^{t-1}$ as the initial model to perform multiple updates on its local dataset $\mathcal{D}_{k}$ and gets the new model $\boldsymbol{\theta}^{t}_{k}$. Then the clients will only send the accumulated gradients $\boldsymbol{g}^{t}_{k} =  \boldsymbol{\theta}^{t-1} - \boldsymbol{\theta}^{t}_{k}$ back to the server, the server aggregates the collected local gradients and updates the global model as the following:
\begin{equation}
\label{eq: fedavg}
    \boldsymbol{\theta}^{t} = \boldsymbol{\theta}^{t-1} - \eta_{s} \frac{1}{|\mathcal{C}^{t}|} \sum\limits_{k \in \mathcal{C}^{t}}  \frac{|\mathcal{D}_{k}|}{\sum_{i\in \mathcal{C}^{t}}|\mathcal{D}_{i}|} \boldsymbol{g}_{k}^{t},
\end{equation}
where $\eta_{s}$ is the server learning rate, $|\mathcal{C}^{t}|$ is the number of clients participating in the current round.

In this paper, we study the optimization problem of FL in the setting where the global data distribution $\mathcal{D}=\bigcup_{k} \mathcal{D}_{k}$ is long-tailed. Previous studies in centralized long-tailed learning~\cite{decoupling} propose to decouple the training on the long-tailed classification tasks into the representation learning and classifier learning phases, and point out that performing class-level re-weighting rather than instance-balanced training brings negative impact on the representation learning, and the imbalanced data distribution mainly affects the classifier learning. Thus, \textbf{our main motivation is to effectively re-balance the classifier when dealing with the long-tailed global data.} Specifically, in order to tackle the problem of the lack of a global balanced dataset in FL, we manage to make each client re-balance the classifier locally during training, by taking great advantage of the abundant real data stored in the local datasets.

\subsection{Our Optimization Target}
\label{subsec: our optimization target}
\begin{figure*}[!t]
	\centering
	\includegraphics[width=\linewidth]{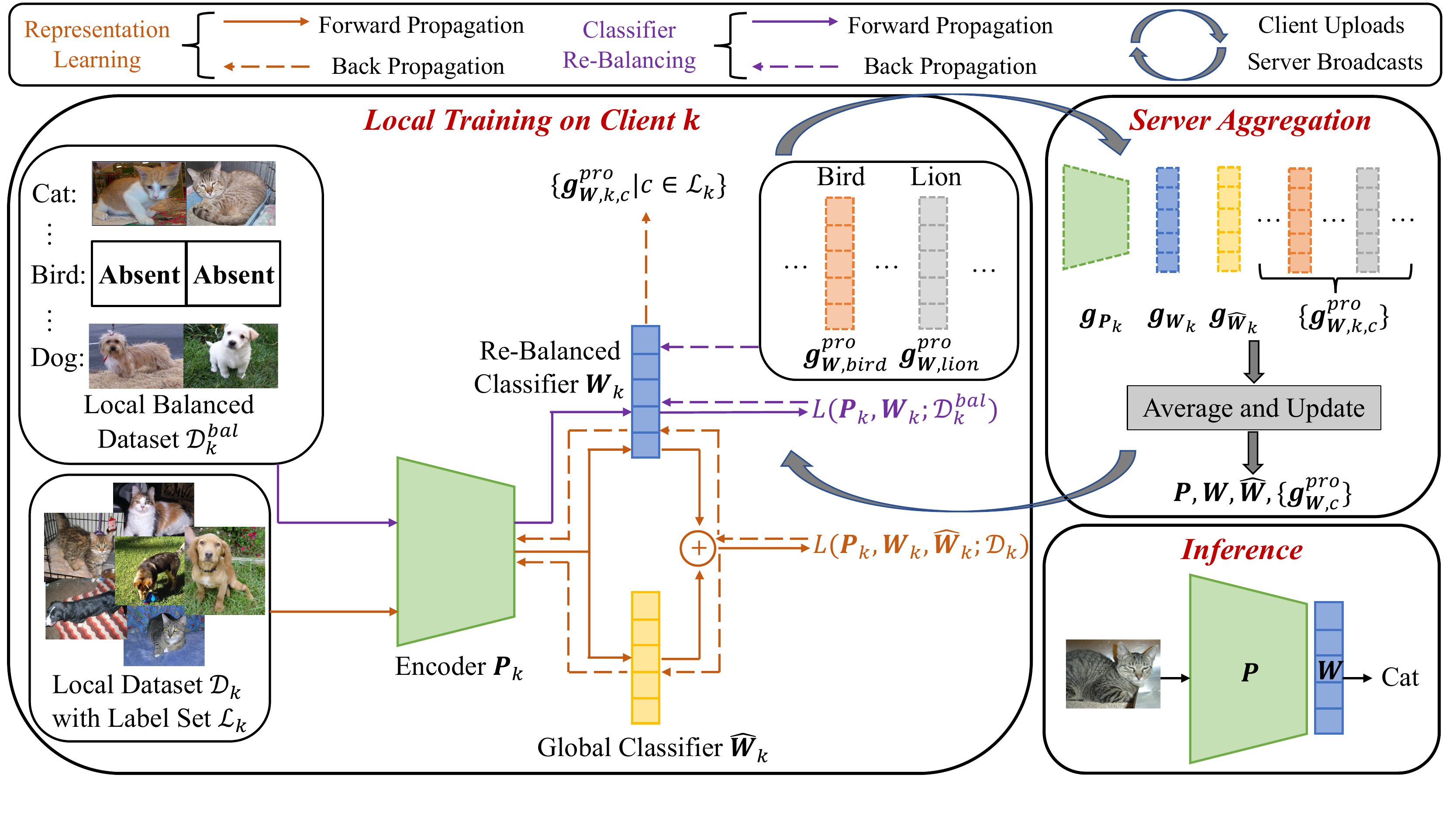}
	\caption{The full illustration of our method. We add a new global classifier $\widehat{\boldsymbol{W}}$, and perform the instance-balanced training on the whole network (including the encoder and the two classifiers). Furthermore, we propose a novel algorithm to re-balance the original classifier during the local training by integrating the local real data and the global gradient prototypes ($\boldsymbol{g}_{\boldsymbol{W},c}^{pro}$ in the figure) to form a local balanced dataset for adjusting the original classifier. In the inference phase, we only keep the global encoder $\boldsymbol{P}$ and the re-balanced classifier $\boldsymbol{W}$.}
	\label{fig: algorithm}
\end{figure*}

We split the original model architecture $\boldsymbol{\theta}$ into two parts: the representation encoder $\boldsymbol{P}$ and the classifier $\boldsymbol{W}$, and aim to re-balance $\boldsymbol{W}$ during the local training to make it behave well on the class-balanced data distribution $\mathcal{D}^{bal}$. However, re-balancing classifier during (instead of after) the representation learning phase leads to a contradictory optimization target:
\begin{equation}
\label{eq: wrong targets}
\begin{aligned}
(\boldsymbol{P}^{*}, \boldsymbol{W}^{*}) &=
\mathop{\arg\min}\limits_{\boldsymbol{P}, \boldsymbol{W}} L(\boldsymbol{P}, \boldsymbol{W};\bigcup_{k} \mathcal{D}_{k}) \\ 
&=
\mathop{\arg\min}\limits_{\boldsymbol{P}, \boldsymbol{W}}  \sum_{k=1}^{N} \frac{|\mathcal{D}_{k}|}{\sum_{i=1}^{N}|\mathcal{D}_{i}|}L(\boldsymbol{P}, \boldsymbol{W};\mathcal{D}_{k}), \\
\text{s.t.} \qquad
\boldsymbol{W}^{*} &= \mathop{\arg\min}\limits_{\boldsymbol{W}} L(\boldsymbol{P}^{*},\boldsymbol{W}; \mathcal{D}^{bal}),
\end{aligned}
\end{equation}
As we can see, when the global data distribution $\mathcal{D}=\bigcup_{k} \mathcal{D}_{k}$ is long-tailed, the above optimization problem has no solution, caused by the contradictory goals when updating $\boldsymbol{W}$. To address to negative impact of performing local classifier re-balancing, we design an architecture of the two-stream classifiers by adding a new classifier $\widehat{\boldsymbol{W}}$ in the training phase, in order to help model the global data distribution $\mathcal{D}$ and make re-balancing $\boldsymbol{W}$ possible. The full illustrations of our model architecture and training process are in Figure~\ref{fig: algorithm}, and we re-formulate our global optimization target as:
\begin{equation}
\label{eq: two targets}
\resizebox{.89\hsize}{!}{$
\begin{aligned}
(\boldsymbol{P}^{*}, \boldsymbol{W}^{*},\widehat{\boldsymbol{W}^{*}}) & = (\boldsymbol{\theta}^{*}, \widehat{\boldsymbol{W}^{*}}) 
\\ &=
\mathop{\arg\min}\limits_{\boldsymbol{\theta},\widehat{\boldsymbol{W}}}  \sum_{k=1}^{N} \frac{|\mathcal{D}_{k}|}{\sum_{i=1}^{N}|\mathcal{D}_{i}|}L(\boldsymbol{\theta},\widehat{\boldsymbol{W}};\mathcal{D}_{k}), \\
\text{s.t.} \qquad
\boldsymbol{W}^{*} &= \mathop{\arg\min}\limits_{\boldsymbol{W}} L(\boldsymbol{P}^{*},\boldsymbol{W}; \mathcal{D}^{bal}).
\end{aligned}
  $}
\end{equation}
By making the combination of two classifiers model the global data distribution in the first part of Eq.~(\ref{eq: two targets}), we make sure that the representation encoder is trained under the instance-balanced training paradigm, which benefits the representation learning most. In the following, we introduce our algorithm to solve Eq.~(\ref{eq: two targets}) from three aspects, including the local training stage, the server aggregation stage, and the inference stage.

\subsection{Classifier Re-Balancing by Integrating Local Real Data with Global Gradient Prototypes}

\begin{algorithm}[t]
    \caption{Local Training Process of RedGrape}
    \label{alg: local training}
    \textbf{Input}: Round number $t$, local data $\mathcal{D}_{k}$ with local label set $\mathcal{L}_{k}$, global model $(\boldsymbol{P}^{t-1}, \boldsymbol{W}^{t-1},\widehat{\boldsymbol{W}}^{t-1})$, global gradient prototypes $\{ \boldsymbol{g}_{\boldsymbol{W}^{t-2},c}^{pro} |c\in \mathcal{L}\}$. 
    
    \begin{algorithmic}[1] 
        \STATE{Calculate local gradient prototypes $\{ \boldsymbol{g}_{\boldsymbol{W}^{t},k,c}^{pro} |c\in \mathcal{L}_{k}\}$ based on Eq.~(\ref{eq: gradient prototypes of last round}).}
        \FOR{Local step $i=1,2,\cdots,I$}
        \STATE Update $\boldsymbol{P}$ and $\widehat{\boldsymbol{W}}$ based on Eq.~(\ref{eq: updating W_hat and P}).
        \STATE Compute batch gradients for $\boldsymbol{W}$ based on Eq.~(\ref{eq: first part of W}).
        \STATE Compute re-balancing gradients for $\boldsymbol{W}$ based on Eq.~(\ref{eq: gradient prototypes}) and Eq.~(\ref{eq: mixed balanced gradients}).
        \STATE Update $\boldsymbol{W}$ based on Eq.~(\ref{eq: updating W}).
        \ENDFOR
        \STATE \textbf{return} Local gradients $(\boldsymbol{P}^{t}_{k}-\boldsymbol{P}^{t-1}, \boldsymbol{W}^{t}_{k}-\boldsymbol{W}^{t-1},\widehat{\boldsymbol{W}}^{t}_{k}-\widehat{\boldsymbol{W}}^{t-1})$, local gradient prototypes $\{ \boldsymbol{g}_{\boldsymbol{W}^{t-1},k,c}^{pro} |c\in \mathcal{L}_{k}\}$. 
    \end{algorithmic}
\end{algorithm}

\subsubsection*{Local Training Stage}
In the local training, each client aims to solve the sub-problem of Eq.~(\ref{eq: two targets}) as
\begin{equation}
\label{eq: local target}
\begin{aligned}
(\boldsymbol{P}^{*}, \boldsymbol{W}^{*},\widehat{\boldsymbol{W}^{*}})
 & =
\mathop{\arg\min}\limits_{\boldsymbol{P},\boldsymbol{W},\widehat{\boldsymbol{W}}} L(\boldsymbol{P},\boldsymbol{W},\widehat{\boldsymbol{W}};\mathcal{D}_{k}), \\
\text{s.t.} \qquad
\boldsymbol{W}^{*} &= \mathop{\arg\min}\limits_{\boldsymbol{W}} L(\boldsymbol{P}^{*},\boldsymbol{W}; \mathcal{D}^{bal}).
\end{aligned}
\end{equation}
It is a constrained optimization problem that is non-trivial to solve, we choose to address it by considering it as a multi-target learning task and optimizing all parameters concurrently. We briefly summarize the whole process of local training in our method in Algorithm~\ref{alg: local training}. To specific, the encoder parameters $\boldsymbol{P}$ and the additional classifier $\widehat{\boldsymbol{W}}$ will be trained under an instance-balanced manner (Line 3). When updating $\boldsymbol{W}$, besides the gradients of the batch samples from $\mathcal{D}_{k}$ (Line 4), our method creates a local balanced dataset $\mathcal{D}^{bal}_{k}$ to help re-balancing $\boldsymbol{W}$ (Line 5-6) following the second target of Eq.~(\ref{eq: local target}). The detailed steps including the following parts:

\paragraph{Updating $\boldsymbol{P}$ and $\widehat{\boldsymbol{W}}$.} In the $t$-th round, 
we perform the normal stochastic gradient decent mechanism\footnote{We do not have the assumption about the local optimizer, and any local optimizer (e.g., SGDM or Adam~\cite{adam}) is acceptable. Here, we take SGD as an example.} in which an instance-balanced dataloader is applied to update $\boldsymbol{P}$ and $\widehat{\boldsymbol{W}}$.\footnote{For simplicity, we omit the bias term here, while our method is still applicable when the bias term exists.} That is, 
for the local step $i=1,2,\cdots,I$, a random batch of examples $\mathcal{B}_{k}^{i}$ is sampled from $\mathcal{D}_{k}$ to perform that:
\begin{equation}
\label{eq: updating W_hat and P}
\resizebox{.89\hsize}{!}{$
\begin{aligned}
\boldsymbol{P}^{i}_{k} &= \boldsymbol{P}^{i-1}_{k} - \eta_{l} \nabla_{\boldsymbol{P}^{i-1}_{k}}L(\boldsymbol{P}^{i-1}_{k},\boldsymbol{W}^{i-1}_{k},\widehat{\boldsymbol{W}}^{i-1}_{k};\mathcal{B}_{k}^{i}),
\\
\widehat{\boldsymbol{W}}^{i}_{k} &= \widehat{\boldsymbol{W}}^{i-1}_{k} - \eta_{l} \nabla_{\widehat{\boldsymbol{W}}^{i-1}_{k}}L(\boldsymbol{P}^{i-1}_{k},\boldsymbol{W}^{i-1}_{k},\widehat{\boldsymbol{W}}^{i-1}_{k};\mathcal{B}_{k}^{i}),
\end{aligned}
 $}
\end{equation}
in which the initial model $(\boldsymbol{P}^{0}_{k}, \boldsymbol{W}^{0}_{k},\widehat{\boldsymbol{W}}^{0}_{k}) $ is chosen as the global model $(\boldsymbol{P}^{t-1}, \boldsymbol{W}^{t-1},\widehat{\boldsymbol{W}}^{t-1})$ received from the server in the current round, $\eta_{l}$ is the local learning rate. One important thing is, when calculating the above loss $l$ on each sample $(\boldsymbol{x},y)$, the representation vector $\boldsymbol{h}:=f(\boldsymbol{x};\boldsymbol{P})$ will be first fed into both two classifiers and get two logits $\boldsymbol{W}^{T}\boldsymbol{h}$ and $\widehat{\boldsymbol{W}}^{T}\boldsymbol{h}$. Then we perform the element-wise addition to get the final logits $\boldsymbol{z}=\boldsymbol{W}^{T}\boldsymbol{h} + \widehat{\boldsymbol{W}}^{T}\boldsymbol{h}$, and use $\boldsymbol{z}$ for the loss calculation. 

\paragraph{Updating $\boldsymbol{W}$.} When updating $\boldsymbol{W}$, one part of gradients comes from the above back propagation process on $\mathcal{B}_{k}^{i}$ as 
\begin{equation}
\label{eq: first part of W}
\boldsymbol{g}_{\boldsymbol{W}^{i-1}_{k}}^{local} = \nabla_{\boldsymbol{W}^{i-1}_{k}}L(\boldsymbol{P}^{i-1}_{k},\boldsymbol{W}^{i-1}_{k},\widehat{\boldsymbol{W}}^{i-1}_{k};\mathcal{B}_{k}^{i}),
\end{equation}
which corresponds to the first part in Eq.~(\ref{eq: local target}). For the second part of Eq.~(\ref{eq: local target}), it needs to calculate the gradients of $\boldsymbol{W}^{i-1}_{k}$ on a small balanced set $\mathcal{D}^{bal}_{k}$, that is supposed to be created in the local. However, there exists difficulty in constructing $\mathcal{D}^{bal}_{k}$ from $\mathcal{D}_{k}$, since it is very likely that there are some classes missing in the local label set $\mathcal{L}_{k}$ of $\mathcal{D}_{k}$ due to the non-i.i.d.\ data partitions. Then, we propose a \textit{mixed gradient re-balancing mechanism} to overcome this challenge by integ\textbf{R}ating local r\textbf{E}al \textbf{D}ata with \textbf{G}lobal g\textbf{RA}dient prototy\textbf{PE}s (\textbf{RedGrape} as our method). 
Specifically, for each class $c$, \textbf{(1)} if the sample quantity of class $c$ in $\mathcal{D}_{k}$ reaches a threshold $T$, we think client $k$ have sufficient samples of class $c$ in its local dataset, and randomly sample $T$ training samples of class $c$ to form $\mathcal{D}^{bal}_{k,c}$ for $\mathcal{D}^{bal}_{k}$.\footnote{In different rounds, client $k$ can choose different $T$ samples of class $c$ for $\mathcal{D}^{bal}_{k,c}$, in order to make fully use of the local real data.} Then, the gradients contributed by class $c$ in $\mathcal{D}^{bal}_{k}$ is
\begin{equation}
\label{eq: real data gradient}
\begin{aligned}
\boldsymbol{g}_{\boldsymbol{W}^{i-1}_{k},c} = \nabla_{\boldsymbol{W}^{i-1}_{k}}L(\boldsymbol{P}^{i-1}_{k},\boldsymbol{W}^{i-1}_{k};D^{bal}_{k,c}).
\end{aligned}
\end{equation} 
\textbf{(2)} If client $k$ does not have enough data of class $c$ in its local dataset, we choose to estimate the gradient contribution of class $c$ with the global gradient prototype of class $c$ in the $(t-1)$-th round, which is the averaged gradient of training samples belonging to class $c$ w.r.t.\ the classifier $\boldsymbol{W}^{t-2}$ across all available clients in the last round~\cite{creff}:
\begin{equation}
\label{eq: gradient prototypes}
\begin{aligned}
\boldsymbol{g}_{\boldsymbol{W}^{t-2},c}^{pro} = \frac{1}{|\mathcal{C}^{t-1}_{c}|}  \sum_{k \in \mathcal{C}^{t-1}_{c}}\boldsymbol{g}^{pro}_{\boldsymbol{W}^{t-2},k,c},
\end{aligned}
\end{equation} 
\begin{equation}
\label{eq: gradient prototypes of last round}
\begin{aligned}
\boldsymbol{g}_{\boldsymbol{W}^{t-2},k,c}^{pro} = \nabla_{\boldsymbol{W}^{t-2}}L(\boldsymbol{P}^{t-2},\boldsymbol{W}^{t-2};\mathcal{D}_{k,c}) ,
\end{aligned}
\end{equation} 
where $\mathcal{C}_{c}^{t-1}$ represents the set of clients sampled in the $(t-1)$-th round and have the training samples of class $c$, and $\mathcal{D}_{k,c}$ denotes all training samples of class $c$ in $\mathcal{D}_{k}$. Thus, it requires each client sampled in the previous round to first calculate the local gradient prototype of each class $c \in \mathcal{L}_{k}$ on the same model $(\boldsymbol{P}^{t-2},\boldsymbol{W}^{t-2})$, return $\{ \boldsymbol{g}_{\boldsymbol{W}^{t-2},k,c}^{pro} | c\in \mathcal{L}_{k}\}$ back to the server along with other local gradients, and receive the global gradient prototypes averaged and sent by the server.
Then, the final gradients on the local balanced dataset to optimize the second part of Eq.~(\ref{eq: local target}) is 
\begin{equation}
\label{eq: mixed balanced gradients}
\begin{aligned}
\boldsymbol{g}_{\boldsymbol{W}^{i-1}_{k}}^{bal} = \frac{1}{|\mathcal{L} |} (\sum\limits_{c \in \mathcal{L}_{k}^{bal}} \boldsymbol{g}_{\boldsymbol{W}_{k}^{i-1},c} + \sum\limits_{c \in \mathcal{L} \backslash \mathcal{L}_{k}^{bal} } \boldsymbol{g}_{\boldsymbol{W}^{t-2},c}^{pro}) ,
\end{aligned}
\end{equation} 
where $\mathcal{L}^{bal}_{k} \subset \mathcal{L}_{k}$ is the label set in which each classe consists of more than $T$ samples and $\mathcal{L}$ is the entire label set. Finally, the updating rule for $\boldsymbol{W}^{i-1}_{k}$ is\footnote{The local classifier re-balancing starts from the 2nd round.} 
\begin{equation}
\label{eq: updating W}
\resizebox{.89\hsize}{!}{$
\begin{aligned}
\boldsymbol{W}^{i}_{k} = \boldsymbol{W}^{i-1}_{k} - \eta_{l} &\left[\boldsymbol{g}_{\boldsymbol{W}^{i-1}_{k}}^{local}+ \lambda \boldsymbol{g}_{\boldsymbol{W}^{i-1}_{k}}^{bal} \frac{\| \boldsymbol{g}_{\boldsymbol{W}^{i-1}_{k}}^{local} \| }{\| \boldsymbol{g}_{\boldsymbol{W}^{i-1}_{k}}^{bal} \|} \right],
\end{aligned}
 $}
\end{equation} 
where $\lambda$ is a re-balance factor to control the re-balancing strength for updating $\boldsymbol{W}$. In Eq.~(\ref{eq: updating W}), we normalize the scale\footnote{Here, $\| \cdot \|$ represents the Frobenius Norm.} of $\boldsymbol{g}_{\boldsymbol{W}^{i-1}_{k}}^{bal} $ at each step in order to address the unstable training caused by the constant part of $\boldsymbol{g}_{\boldsymbol{W}^{t-1},c}^{pro}$, by making its scale consistent with the decreasing trend of the scale of real gradients during training. 

After training, the new model is $(\boldsymbol{P}^{t}_{k}, 
\boldsymbol{W}^{t}_{k},\widehat{\boldsymbol{W}}^{t}_{k})$, and client $k$ sends the local gradients $(\boldsymbol{g}_{\boldsymbol{P}^{t-1},k}, \boldsymbol{g}_{\boldsymbol{W}^{t-1},k},\boldsymbol{g}_{\widehat{\boldsymbol{W}}^{t-1},k})=(\boldsymbol{P}^{t}_{k}-\boldsymbol{P}^{t-1}, 
\boldsymbol{W}^{t}_{k}-\boldsymbol{W}^{t-1},\widehat{\boldsymbol{W}}^{t}_{k}-\widehat{\boldsymbol{W}}^{t-1})$ 
along with the local gradient prototypes $\{ \boldsymbol{g}_{\boldsymbol{W}^{t-1},k,c}^{pro} | c\in  \mathcal{L}_{k}\}$ to the server.

\subsubsection{Server Aggregation Stage}
The server first aggregates the gradients and updates the global model as 
\begin{equation}
\label{eq: server aggregation}
\begin{aligned}
& (\boldsymbol{P}^{t},\boldsymbol{W}^{t}, \widehat{\boldsymbol{W}}^{t}) = (\boldsymbol{P}^{t-1},\boldsymbol{W}^{t-1}, \widehat{\boldsymbol{W}}^{t-1}) \\ & - \eta_{s} \sum\limits_{k \in \mathcal{C}^{t}} {\frac{|\mathcal{D}_{k}|}{\sum_{i\in \mathcal{C}^{t}}|\mathcal{D}_{i}|}} 
(\boldsymbol{g}_{\boldsymbol{P}^{t-1},k}, \boldsymbol{g}_{\boldsymbol{W}^{t-1},k},\boldsymbol{g}_{\widehat{\boldsymbol{W}}^{t-1},k}).
\end{aligned}
\end{equation} 
Also, the server needs to update the global gradient prototypes as
\begin{equation}
\label{eq: update global gradient prototypes}
\boldsymbol{g}_{\boldsymbol{W}^{t-1},c}^{pro} = 
\left\{
\begin{aligned}
& \frac{1}{|\mathcal{C}^{t}_{c}|}  \sum_{k \in \mathcal{C}^{t}_{c}}\boldsymbol{g}^{pro}_{\boldsymbol{W}^{t-1},k,c},\quad \mathcal{C}_{c}^{t} \neq \emptyset,\\
& \boldsymbol{g}_{\boldsymbol{W}^{t-2},c}^{pro},\quad \mathcal{C}_{c}^{t} = \emptyset,
\end{aligned}
\right.
\end{equation}
in which the second case corresponds to the situation where all clients in $C^{t}$ from the current round do not contain samples of class $c$. In this case, we re-use the global gradient prototype of class $c$ from the previous round. 
The updated global model and global gradient prototypes are broadcast to the sampled clients in the next round.

\subsection*{Inference Stage}
After federated training, we only keep the re-balanced classifier $\boldsymbol{W}$ and abandon $\widehat{\boldsymbol{W}}$ in the reference stage:
\begin{equation}
\label{eq: inference}
y_{\text{pred}} = \mathop{\arg\max}_{i} [\boldsymbol{W}^{T}f(x;\boldsymbol{P}) ].
\end{equation} 

\section{Experiments and Analysis}
\subsection{Experimental Settings}
\paragraph{Datasets and Models}
We conduct experiments on three popular image classification benchmarks: MNIST~\cite{mnist}, CIFAR-10 and CIFAR-100~\cite{cifar}. We follow existing studies~\cite{ldam,creff} to create the long-tailed versions of training sets of above three datasets (i.e., MNIST-LT, CIFAR-10/100-LT), and keep the test sets as balanced. We first define the term \textit{Imbalance Ratio}: $\text{IR}=\frac{\max_{c}\{ n_{c} \} }{\min_{c} \{n_{c} \} }$, which is the ratio between the maximum sample number across all classes and the minimum sample number across all classes, to reflect the imbalance degree of the global data distribution. Then, the training sample quantity of each class follows an exponential decay. We choose $\text{IR}=10,50,100$ in our main experiments. Furthermore, we follow the existing studies~\cite{fedopt,creff} to adopt the Dirichlet distribution $\text{Dir}(\alpha)$ for the non-i.i.d.\ data partitioning, in which $\alpha$ controls the non-i.i.d.\ degree. We set $\alpha=1.0$ in our main experiments, and put the results of other $\alpha$s in the Appendix. We use the convolutional neural network (CNN)~\cite{fedavg} for MNIST, and use ResNet-56~\cite{resnet} for CIFAR-10/100 datasets. 
More details can be found in the Appendix.

\paragraph{Baseline Methods}
We compare our method with the existing federated long-tailed learning algorithms, including the traditional FedAvg algorithm with the CrossEntropy Loss (FedAvg+CE) applied in the local training~\cite{fedavg}, Fed-Focal Loss~\cite{fed-focal}, Ratio Loss~\cite{ratio}, CLIMB~\cite{climb}, and the state-of-the-art method CReFF~\cite{creff}.

\paragraph{Training Details}
We conduct experiments in two popular FL settings based on the ratio of clients participating in each round: (1) \textbf{Full client participation} setting: all clients participate in updating the global model in each round, and the total number of clients is 10 in this setting; (2) \textbf{Partial client participation} setting: the total number of clients is 50 but only 10 clients are randomly sampled in each round. We adopt SGDM as the optimizer for local training. The local learning rate is 0.01 for MNIST-LT and 0.1 for CIFAR-10/100-LT. The number of local epochs is 5 for all datasets. As for our method, the re-balance factor $\lambda$ is fixed as 0.1 in all experiments, and we explore the effect of different values of $\lambda$ in Section~\ref{subsec: effect of lambda}. The quantity threshold $T$ for each class to create the local balanced dataset is set as 8 for MNIST-LT and CIFAR-10-LT, and 2 for CIFAR-100-LT, and we put further discussion in Section~\ref{subsec: quantity threshold}. Each experiment is run on 3 random seeds. Complete training details (e.g., the number of communication rounds in each setting, detailed hyper-parameters of other baselines) are in the Appendix. 
Our code is implemented on the FedML~\cite{fedml} platform.\footnote{Our code will be released upon acceptance.} 

\subsection{Main Results}

\begin{table*}[t!]
\begin{center}
\sisetup{detect-all,mode=text}
\begin{tabular}{lccccccccc}
\toprule
\multirow{2.5}{*}{\begin{tabular}[c]{@{}l@{}}Method \end{tabular}} &   \multicolumn{3}{c}{MNIST-LT}  & \multicolumn{3}{c}{CIFAR-10-LT} & \multicolumn{3}{c}{CIFAR-100-LT}  \\
\cmidrule(lr){2-4}
\cmidrule(lr){5-7}
\cmidrule(lr){8-10}
&  $\text{IR}=10$   &  $\text{IR}=50$  &   $\text{IR}=100$    & $\text{IR}=10$   &  $\text{IR}=50$  &   $\text{IR}=100$ & $\text{IR}=10$   &  $\text{IR}=50$  &   $\text{IR}=100$  \\
\midrule
FedAvg+CE   & 97.99 & 95.98 & 92.71   &  76.21 &  68.41 & 59.83 &   49.08 & 36.47 & 33.28 \\
Fed-Focal Loss   & 97.90  & 96.14 & 92.97   & 77.92 & 61.21 & 59.86 &  48.14 & 35.51 & 30.05  \\
Ratio Loss  & 97.96 & 96.20 & 92.99 & 78.58 & 68.01 &  59.27 & 48.30 & 37.62 & 31.92  \\
CLIMB  & 97.89 & 95.87 & 92.71  & 78.95 & 66.25 & 57.67 &  49.27 & 36.13 & 32.18  \\
CReFF  & 97.68 &  96.49 & 93.85 & 83.18 & 73.46 & 69.36 &  46.58 & 35.82 & 33.46\\
\midrule
Ours  &  \textbf{98.34} &  \textbf{97.06} & \textbf{95.73} &  \textbf{83.74} & \textbf{74.01} & \textbf{71.04}  & \textbf{51.09} &\textbf{38.49} & \textbf{34.63}  \\
\bottomrule
\end{tabular}
\end{center}
\caption{Results under the \textbf{full client participation} setting. We report the overall test accuracy on the balanced testing set of each dataset.}
\label{tab: main results full}
\end{table*}

\begin{table*}[t!]
\begin{center}
\sisetup{detect-all,mode=text}
\begin{tabular}{lccccccccc}
\toprule
\multirow{2.5}{*}{\begin{tabular}[c]{@{}l@{}}Method \end{tabular}} &   \multicolumn{3}{c}{MNIST-LT}  & \multicolumn{3}{c}{CIFAR-10-LT} & \multicolumn{3}{c}{CIFAR-100-LT}  \\
\cmidrule(lr){2-4}
\cmidrule(lr){5-7}
\cmidrule(lr){8-10}
&  $\text{IR}=10$   &  $\text{IR}=50$  &   $\text{IR}=100$    & $\text{IR}=10$   &  $\text{IR}=50$  &   $\text{IR}=100$ & $\text{IR}=10$   &  $\text{IR}=50$  &   $\text{IR}=100$   \\
\midrule
FedAvg+CE   &  95.51 & 91.82 & 89.92 & 60.38 & 45.15 & 40.06  &  40.81 &  24.62 & 22.08 \\
Fed-Focal Loss   & 96.79  & 92.59 & 90.45   & 61.16 & 46.20 & 41.10 &  40.85 & 24.73 & 20.17  \\
Ratio Loss  & 95.17 & 91.10 & 89.64 & 63.97 & 44.22 &  42.11 & 40.96 & 24.12 & 23.06  \\
CLIMB  & 95.67 & 92.24 & 89.75  & 61.75 & 46.91 & 42.02 &  40.64 & 23.99 & 21.44  \\
CReFF  & 96.29 &  94.16 & 92.16 & 69.38 & 60.52 & 55.63 &  39.38 & 25.42 & 24.77\\
\midrule
Ours  &  \textbf{97.54} &  \textbf{95.17} & \textbf{93.61} &  \textbf{71.68} & \textbf{61.42} & \textbf{57.11}  & \textbf{42.97} &\textbf{27.73} & \textbf{25.64}  \\
\bottomrule
\end{tabular}
\end{center}
\caption{Results under the \textbf{partial client participation} setting. We report the overall test accuracy on the balanced testing set of each dataset.}
\label{tab: main results partial}
\end{table*}

In the main paper, we report the averaged accuracy over the last 10 rounds on the balanced testing set of each dataset following existing studies~\cite{fedopt}. We also display the averaged test accuracy on tail classes in each setting in the Appendix to show that our method can significantly bring improvement to the model's performance on tail classes. The results under the full client participation setting are in Table~\ref{tab: main results full}, and Table~\ref{tab: main results partial} displays the results under the partial client participation setting. We can draw the main conclusion from these tables as: \textbf{our method can consistently outperform the existing algorithms in all settings.}

As we can see, Fed-Focal Loss achieves lower performance than the FedAvg with CE loss in some settings, which validates the claim that directly applying the centralized long-tailed learning methods can not help to address the global class imbalance problem in FL, as it ignores the mismatch between the global and the local imbalance patterns. Ratio Loss and CLIMB apply the class-level re-weighting and client-level re-weighting idea separately, and gain slight improvement compared with FedAvg. We analyze that the reason for the limited improvement lies in that though the re-weighting practice helps the model to focus more on the learning of tail classes, it is not conducive to the representation learning on the abundant data of the head classes~\cite{decoupling}. Moreover, the assumption of obtaining a global auxiliary dataset makes Ratio Loss impractical in real cases. 

\begin{figure}[!t]
	\centering
	\includegraphics[width=\linewidth]{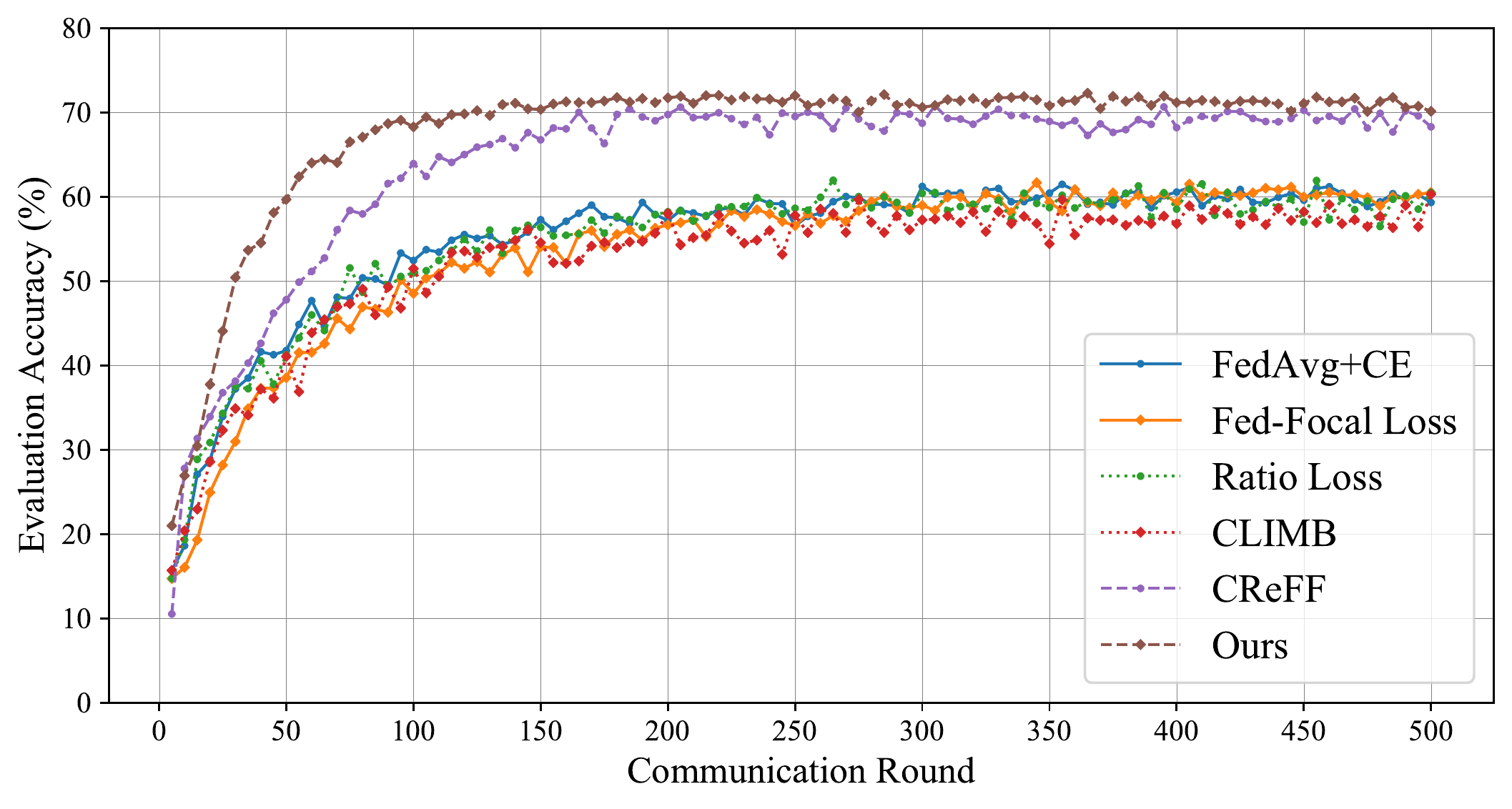}
	\caption{The test accuracy curves on CIFAR-10-LT with $\text{IR}=100$ under the full client participation setting. Our method achieves faster convergence speed and better performance than all existing baselines.}
	\label{fig: convergence speed}
\end{figure}

The superior performance of CReFF helps to validates the effectiveness of the classifier re-balancing on the final performance. However, the optimization of the federated features requires massive computations on the server (especially when the number of classes is large), and the federated features from the same class may converge to be similar. Thus, the re-trained classifier faces the problem that it may overfit on the highly similar and small amount of the federated features (reflected in the poorer performance under smaller $\text{IR}$). Our method instead takes full advantage of the local real data integrated with the global gradient prototypes to locally re-balance the classifier while maintaining the good effects of instance-balanced training on the representation learning, and consistently outperforms all previous methods by a large margin. Compared with CReFF and Ratio Loss, we do not have extra requirements except for the aggregation procedures on the server, and produce a re-balanced classifier that has better generalization ability with the help of abundant real data.

We further display the evaluation accuracy curve after each round in CIFAR-10-LT ($\text{IR}=100$) under the full client participation setting in Figure~\ref{fig: convergence speed}. As we can see, \textbf{our method not only has the best converged performance, but also achieves much faster convergence speed than all baseline methods.} That is because our method re-balances the classifier at each local training step, and this makes it converge faster to the optimal balanced classifier.

\subsection{Results in Another Class Imbalance Setting}
\label{subsec: binary class imbalance}

\begin{table}[t!]
\small
\begin{center}
\setlength{\tabcolsep}{5.0pt}
\sisetup{detect-all,mode=text}
\begin{tabular}{lcccc}
\toprule
\multirow{2.5}{*}{\begin{tabular}[c]{@{}l@{}}Method \end{tabular}} &   \multicolumn{2}{c}{Full Participation}  & \multicolumn{2}{c}{Partial Participation}  \\
\cmidrule(lr){2-3}
\cmidrule(lr){4-5}
&  MNIST   &  CIFAR-10 &   MNIST   &  CIFAR-10  \\
\midrule
FedAvg+CE   & 93.91 & 70.91 & 91.97 & 61.27\\
Fed-Focal Loss   & 93.53 & 70.77 & 91.77 & 59.18\\
Ratio Loss  & 94.23 & 73.22 & 92.32 & 62.86  \\
CLIMB  & 93.89 & 72.04 & 92.11 & 63.18\\
CReFF  & 96.70 & 78.98 & 96.01 & 71.41\\
\midrule
Ours  & \textbf{96.86} & \textbf{79.88} & \textbf{96.43} & \textbf{73.31} \\
\bottomrule
\end{tabular}
\end{center}
\caption{Results in the binary class imbalance setting. $\text{IR}=100$ and the randomly chosen tail classes are 0, 7 and 8.}
\label{tab: binary class imbalance results}
\end{table}

We also conduct experiments in a binary class imbalance setting in FL~\cite{climb}, in which three classes are randomly chosen as the tail classes, and they are assigned with $1/\text{IR}$ number of sampled compared with other normal/head classes. The experiments are conducted on MNIST and CIFAR-10 datasets with $\text{IR}=100$, and other experimental settings are kept as the same as that in our main experiments. The results are in Table~\ref{tab: binary class imbalance results}. The conclusion remains the same that, \textbf{our method achieves the best performance in all cases.}   

\section{Further Explorations}

In this section, we make further explorations about the two crucial hyper-parameters of our method. The following experiments are conducted on CIFAR-10-LT with $\text{IR}=100$ under the full client participation setting.

\subsection{Re-Balancing Strength Decides on The Convergence Trade-off}
\label{subsec: effect of lambda}

\begin{figure}[!t]
	\centering
	\includegraphics[width=\linewidth]{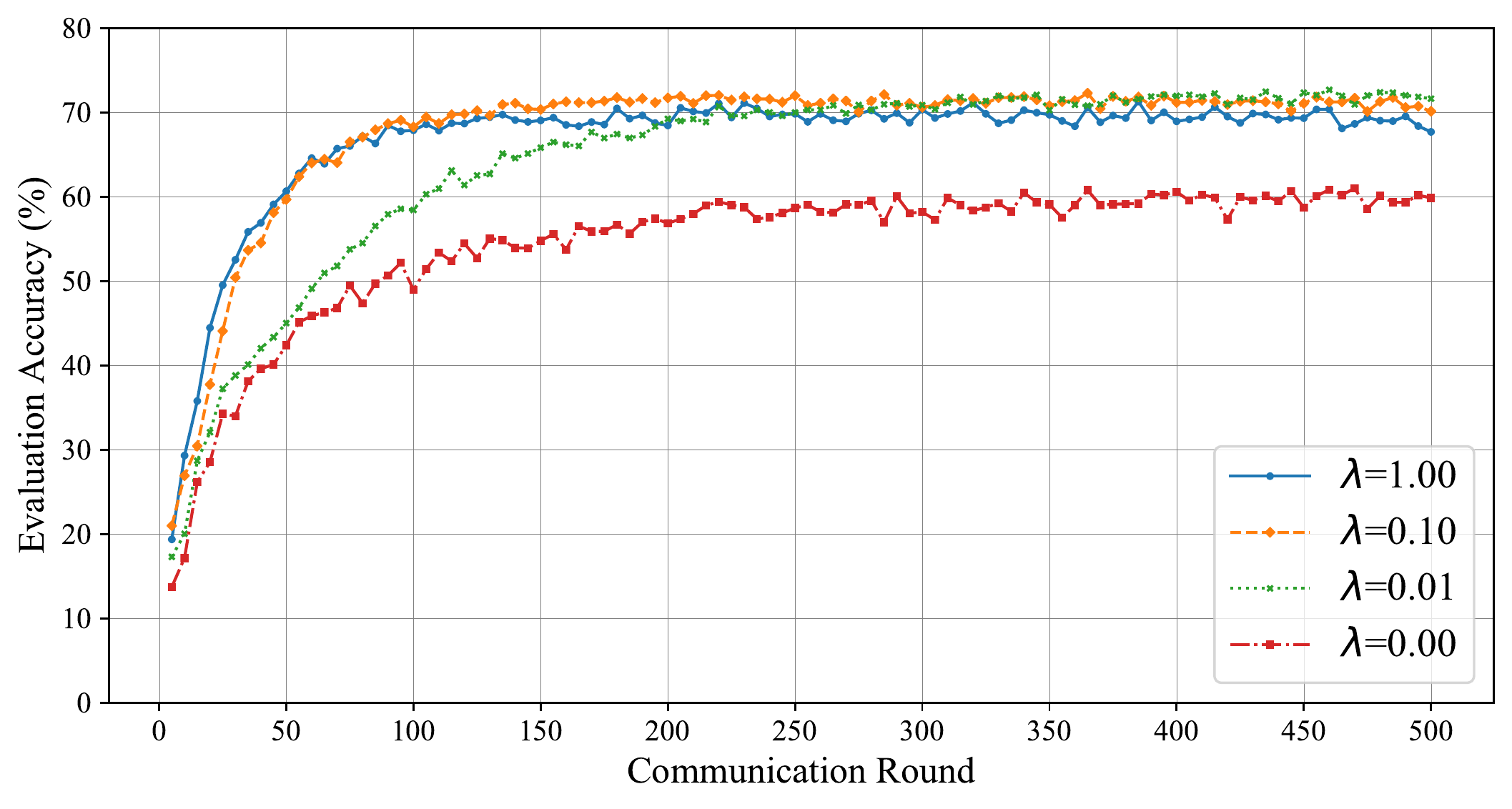}
	\caption{The test accuracy of using different re-balance factor $\lambda$ in CIFAR-10-LT. Smaller $\lambda$ leads to slower convergence speed but relatively better generalization ability.}
	\label{fig: effect of lambda}
\end{figure}

In order to solve the optimization target of Eq.~(\ref{eq: two targets}), we consider updating $\boldsymbol{W}$ in a multi-task learning setting as Eq.~(\ref{eq: updating W}), where a re-balance factor $\lambda$ is used to control the re-balancing strength. Here, we conduct experiments to explore the effect of different $\lambda$s on the model's performance, and the results are shown in Figure~\ref{fig: effect of lambda}. We find that the smaller $\lambda$ results in slower convergence speed but obtains relatively better performance of the converged model. We analyze the reason lies in that, the $\boldsymbol{g}_{\boldsymbol{W}^{i-1}_{k}}^{bal}$ contains a part of global gradient prototypes calculated in the previous round and is a constant when updating $\boldsymbol{W}$. It will adversely affect the model's convergence in the late stage of the training when we are still using a large $\lambda$ to re-balance the classifier. An interesting direction to improve our method is designing an adaptive $\lambda$ that decays along with the training, which we leave to future work. When $\lambda=0.0$, the addition of two classifiers equals to one normal classifier used in FedAvg, so FedAvg is a special case of our method in this case. 

\subsection{Local Real Data Plays An Important Role in Re-Balancing The Classifier}
\label{subsec: quantity threshold}
\begin{figure}[!t]
	\centering
	\includegraphics[width=\linewidth]{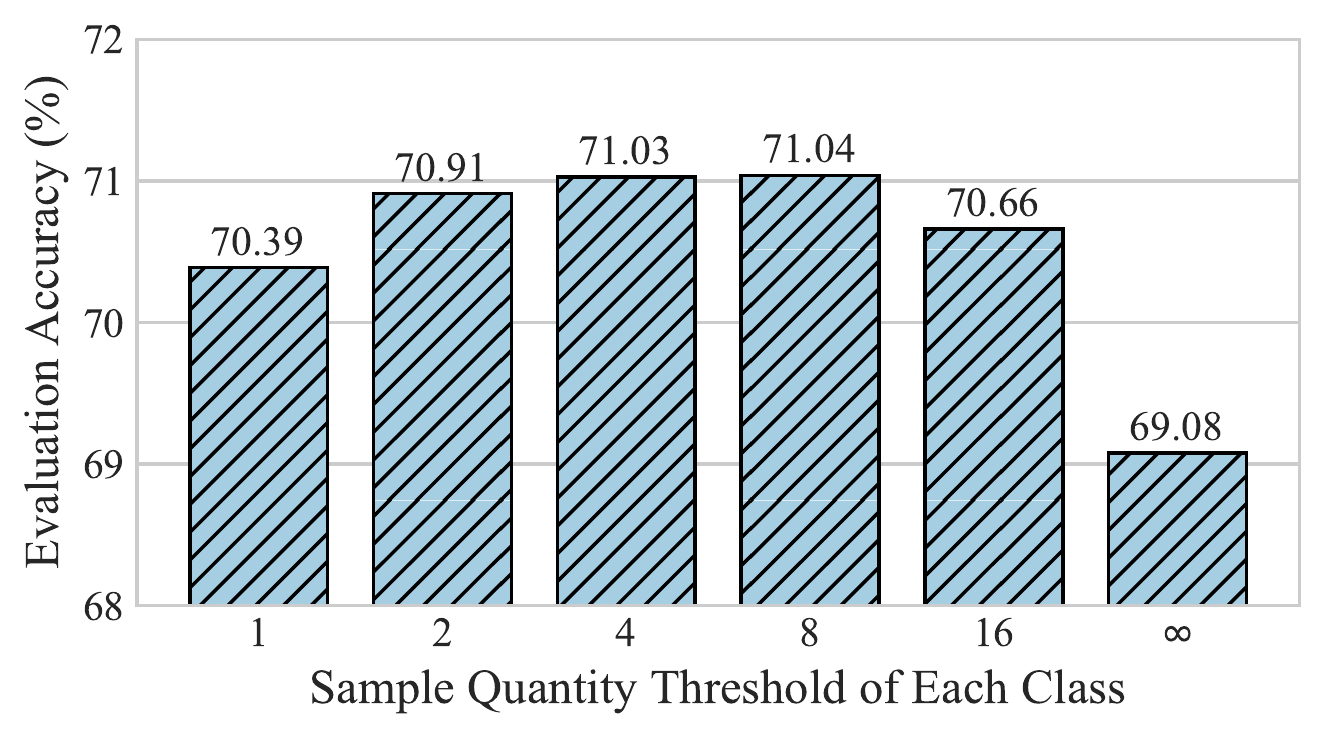}
	\caption{The test accuracy of using different sample quantity thresholds for each class when creating the local balanced datasets in CIFAR-10-LT.}
	\label{fig: quantity threshold}
\end{figure}

During creating the local balanced datasets, we set a threshold $T$ to decide whether the local clients own the enough data of a specific class. Larger $T$ decreases the number of available classes in which the real data can be used to calculate the gradients for classifier re-balancing, while smaller $T$ leads to the relatively unreliable gradients of class $c$. We then explore the effect of different $T$s, and put the results in Figure~\ref{fig: quantity threshold}. We indeed observe a trade-off pattern as expected and find that $T=4,8$ are the most proper choices. $T=\infty$ means we remove the role of local real data on the classifier re-balancing and only use the global gradient prototypes instead, and we find the performance degrades greatly, which verifies the large benefits of using local real data to adjust the classifier.

\section{Conclusion}
In this paper, motivated by the decoupled training idea, we propose a novel and effective classifier re-balancing algorithm for tackling federated long-tailed learning. 
In order to overcome the lack of a public balanced dataset in FL, we propose to re-balance the classifier during local training by integrating local real data with global gradient prototypes. Furthermore, in order to address the problem of contradictory optimization goals during training brought by performing local classifier re-balancing, we introduce a two-stream classifiers architecture to help model the global data distribution. Thorough experiments verify the great effectiveness of our method over strong baselines without extra data requirements.

\section*{Ethical Statement}
Our purpose is to address the optimization problem of FL on the non-i.i.d.\ and long-tailed data and help to learn a better global model that has good performance on all classes. The datasets used in our experiments are all publicly available. Also, our method only requires the normal gradients transmission between the server and the clients, which will not expose the local data privacy and does not have any ethical concerns.

\section*{Acknowledgments}
This work was supported by a Tencent Research Grant. Xu Sun is the corresponding author of this paper.







\bibliographystyle{named}
\bibliography{ijcai23}

\appendix
\section{Detailed Experimental Settings}
\subsection{Datasets and Models}
Here, we introduce the datasets and the backbone models we used in our experiments. We choose three classical image classification tasks, including MNIST~\cite{mnist}, CIFAR-10 and CIFAR-100~\cite{cifar}. We then follow existing centralized and federated long-tailed learning studies~\cite{ldam,creff} to create the long-tailed versions of the training sets of above datasets (i.e., MNIST-LT, CIFAR-10/100-LT). Specifically, the long-tailed degree is controlled by a ratio called the \textit{Imbalance Ratio}: $\text{IR=}\frac{\max_{c}\{ n_{c} \} }{\min_{c} \{n_{c} \} }$, where $n_{c}$ represents the sample quantity of class $c$ (0-indexed). Then, we manage to make the sample quantity of each class follow an exponential decay trend:
\begin{equation}
    n_{c} = n_{0} \times \left(\frac{1}{\text{IR}} \right)^{\frac{c}{C-1}}, \quad c=0,\cdots, C-1.
\end{equation}
As for the non-i.i.d.\ data partitioning, we follow existing studies~\cite{fedopt,creff} to adopt the Dirichlet distribution $\text{Dir}(\alpha)$. Smaller $\alpha$ means the heavier non-i.i.d.\ degree. We choose $\alpha=1.0$ in our main paper, and we also put the results on different $\alpha$s in Appendix~\ref{appendix: diffrent alpha}.

We use the same convolutional neural network (CNN) used in~\cite{fedavg} for experiments on MNIST-LT, and adopt the ResNet-56~\cite{resnet} as the backbone model for CIFAR-10/100-LT.

\subsection{Complete Training Details}

\subsubsection*{Local Training Settings}

We utilize SGDM as the local optimizer in all experiments. The local learning rate is $0.01$ for MNIST-LT, and $0.1$ for CIFAR-10/100-LT. For all three datasets, the batch size for local training is 64, and the number of local training epochs is 5. As mentioned in main paper, we perform experiments in both full client participation and partial client participation settings. We set different total communication rounds in different setting considering the different convergence speeds of the global models: (1) In the full client participation setting, the number of communication rounds is 200 for MNIST-LT, and 500 for CIFAR-10/100-LT. (2) In the full client participation setting, the number of communication rounds is 500 and 1000 for MNIST-LT and CIFAR-10/100-LT separately.

\subsubsection*{Server Aggregation Settings}
During server aggregation, we follow the same procedure as that in FedAvg to aggregate the collected local gradients in the current round, and update the global model with the averaged gradients. The server learning rate is tuned as $1.0$ for all experiments. Furthermore, as for CReFF and our method, the server needs to update the global gradient prototypes (refer to Section 3.3 in our main paper) by averaging local gradient prototypes. However, compared with CReFF, we do not have extra requirements on the server to make it perform further optimization and training process.

\subsubsection*{Hyper-Parameters of Each Method}
Here, we introduce the choices of hyper-parameters used in each method in detail. 

\paragraph{Fed-Focal Loss:} Fed-Focal Loss~\cite{fed-focal} directly applies Focal Loss~\cite{focal-loss} to the local training. The form of Focal Loss is
\begin{equation}
\label{eq: focal loss}
L_{focal} = -(1 - p_{T})^{\gamma} \log(p_{T}),
\end{equation}
where $p_{T}$ is the predicted probability of the sample corresponding to the ground truth class. We set $\gamma=2$ in our experiments.

\paragraph{Ratio Loss:} Ratio Loss~\cite{ratio} applies the class-level re-weighting practice by first estimating the global imbalance pattern on the server with an auxiliary balanced dataset. Its form can be written as
\begin{equation}
\label{eq: ratio loss}
L_{ratio} = (\alpha + \beta \mathbb{R})L_{CE},
\end{equation}
where $L_{CE}$ is the traditional CrossEntropy Loss, $\mathbb{R}$ is the ratio vector that contains the relatively estimated sample quantity of each class on the server, $\alpha$ and $\beta$ are two hyper-parameters. Thus, we follow the original study~\cite{ratio} to set the sample number of each class on the auxiliary balanced dataset to be 32, $\alpha=1.0$, $\beta=0.1$.

\paragraph{CLIMB:} CLIMB~\cite{climb} aims to perform the client-level re-weighting to up-weight the aggregation weights for the local gradients with larger local training losses, as the global model behaves poorly on these clients' local data. The hyper-parameters in CLIMB includes a tolerance constant $\epsilon$ and a dual step size $\eta_{D}$. In our experiments, we follow the original setting to set $\epsilon=0.01$ for MNIST-LT and $\epsilon=0.1$ for CIFAR-10/100-LT, set $\eta$ as 2.0 and 0.1 for MNIST-LT and CIFAR-10/100-LT separately.

\paragraph{CReFF:} CReFF~\cite{creff} needs to create a set of federated features on the server, of which the number per class is 100. Following the original setting, the optimization steps on the federated features is 100, the classifier re-training steps is 300. Further, the learning rate of optimizing the federated features is 0.1 for all datasets, and the learning rate of classifier re-training is kept as the same as that used in the local training on that dataset.

\paragraph{RedGrape (Ours):} Our method introduces two hyper-parameters: $\lambda$ for the classifier re-balancing strength, and $T$ for the sample quantity threshold of each class on creating local balanced datasets. We put the detailed discussions about these two hyper-parameters in our main paper. The recommended search grids for $\lambda$ are $\{1.0, 0.1, 0.01\}$, and $\{2, 4, 8\}$ for $T$.

\subsubsection*{Code and Infrastructure}
Our code is implemented based on the open-sourced FL platform FedML~\cite{fedml}. We will release our code upon acceptance. Our experiments are conducted on 8 * GeForce RTX 2080 Ti.

\begin{table*}[t!]
\begin{center}
\sisetup{detect-all,mode=text}
\begin{tabular}{lccccccccc}
\toprule
\multirow{2.5}{*}{\begin{tabular}[c]{@{}l@{}}Method \end{tabular}} &   \multicolumn{3}{c}{MNIST-LT}  & \multicolumn{3}{c}{CIFAR-10-LT} & \multicolumn{3}{c}{CIFAR-100-LT}  \\
\cmidrule(lr){2-4}
\cmidrule(lr){5-7}
\cmidrule(lr){8-10}
&  $\text{IR}=10$   &  $\text{IR}=50$  &   $\text{IR}=100$    & $\text{IR}=10$   &  $\text{IR}=50$  &   $\text{IR}=100$ & $\text{IR}=10$   &  $\text{IR}=50$  &   $\text{IR}=100$  \\
\midrule
FedAvg+CE   & 96.05 & 90.23 & 82.21 & 67.87  & 55.26 & 29.24 & 36.25 & 13.37 & \phantom{0}8.50\\
Fed-Focal Loss   & 95.84 & 90.47  & 82.62  & 74.94 & 47.37 & 33.86 & 34.40 & 12.53 & \phantom{0}6.49 \\
Ratio Loss  &  96.04 & 90.76 &83.01 & 71.27 & 55.79 & 33.28 & 34.55 & 15.35 & \phantom{0}7.98 \\
CLIMB  &  95.70 & 89.93 & 82.01 &  73.68 & 55.86 & 30.95 & 35.81 & 13.31 & \phantom{0}8.65\\
CReFF  & 95.98 & 91.98 & 86.62 & 82.87 & 66.10 & 57.03 & 38.18 & \textbf{22.77} & \textbf{18.98} \\
\midrule
Ours  & \textbf{96.78}  & \textbf{92.97} & \textbf{89.59} & \textbf{83.86} & \textbf{69.74} & \textbf{60.41} & \textbf{40.11} & 20.19 & 15.58  \\
\bottomrule
\end{tabular}
\end{center}
\caption{Averaged evaluation accuracy on the tail classes under the \textbf{full client participation} setting in main experiments.}
\label{tab: tail results full}
\end{table*}

\begin{table*}[t!]
\begin{center}
\sisetup{detect-all,mode=text}
\begin{tabular}{lccccccccc}
\toprule
\multirow{2.5}{*}{\begin{tabular}[c]{@{}l@{}}Method \end{tabular}} &   \multicolumn{3}{c}{MNIST-LT}  & \multicolumn{3}{c}{CIFAR-10-LT} & \multicolumn{3}{c}{CIFAR-100-LT}  \\
\cmidrule(lr){2-4}
\cmidrule(lr){5-7}
\cmidrule(lr){8-10}
&  $\text{IR}=10$   &  $\text{IR}=50$  &   $\text{IR}=100$    & $\text{IR}=10$   &  $\text{IR}=50$  &   $\text{IR}=100$ & $\text{IR}=10$   &  $\text{IR}=50$  &   $\text{IR}=100$  \\
\midrule
FedAvg+CE   & 89.75 & 80.05 & 74.35 & 51.99 & 24.03  & \phantom{0}2.64 & 29.65 & \phantom{0}8.55 & \phantom{0}5.74  \\
Fed-Focal Loss   & 92.65 & 82.43 & 75.46 & 49.78 & 27.43 & \phantom{0}5.68 & 29.51 & \phantom{0}8.89 & \phantom{0}4.08 \\
Ratio Loss  & 89.83 & 77.93 & 73.35 & 55.54 & 23.81 & \phantom{0}4.87 & 29.28 & \phantom{0}8.36 & \phantom{0}5.86  \\
CLIMB  & 90.09 & 81.49 & 74.58 & 54.45 & 23.90 & \phantom{0}4.14 & 29.85 & \phantom{0}8.37 & \phantom{0}4.97\\
CReFF  & 92.87 & 88.94 & 83.75 & 74.81 & 62.67 & \textbf{45.30} & 34.38 & 17.46 & \textbf{16.58} \\
\midrule
Ours  & \textbf{94.82} & \textbf{90.06} & \textbf{86.38} & \textbf{75.71} & \textbf{63.58} & 43.92 & \textbf{35.34} & \textbf{17.91} & 13.39  \\
\bottomrule
\end{tabular}
\end{center}
\caption{Averaged evaluation accuracy on the tail classes under the \textbf{partial client participation} setting in main experiments.}
\label{tab: tail results partial}
\end{table*}

\section{Results on The Tail Classes in Main Experiments}
In our main paper, we put the results of the overall accuracy on the balanced testing sets of each method, and we have that \textbf{our method consistently outperform all other methods in all settings.} Here, we put the averaged accuracy on the tail classes of each method. Specifically, we define the tail classes as the last 30\% classes the minimum sample quantity.

The results are in Table~\ref{tab: tail results full} and Table~\ref{tab: tail results partial}. As we can see, \textbf{our method brings significant improvement on the tail classes in most cases.} Also, we find that CReFF tends to achieve better performance on the tail classes when the imbalance degree is larger. However, the performance of CReFF on the overall testing sets is worse than our method according to the results in our main paper. This partly validates our analysis that, re-training the classifier on a set number of federated features on the server can indeed re-balance the classifier to some extent, but it is likely to produce a sub-optimal classifier that overfits on these limited number of pseudo features.

\section{Experiments on Different Non-I.I.D.\ Degrees}
\label{appendix: diffrent alpha}

\begin{table*}[t!]
\begin{center}
\sisetup{detect-all,mode=text}
\begin{tabular}{lccccccccc}
\toprule
\multirow{2.5}{*}{\begin{tabular}[c]{@{}l@{}}Method \end{tabular}} &   \multicolumn{3}{c}{MNIST-LT}  & \multicolumn{3}{c}{CIFAR-10-LT} & \multicolumn{3}{c}{CIFAR-100-LT}  \\
\cmidrule(lr){2-4}
\cmidrule(lr){5-7}
\cmidrule(lr){8-10}
&  $\alpha=0.1$   &  $\alpha=1.0$  &   $\alpha=10.0$    &  $\alpha=0.1$   &  $\alpha=1.0$  &   $\alpha=10.0$   &  $\alpha=0.1$   &  $\alpha=1.0$  &   $\alpha=10.0$   \\
\midrule
FedAvg+CE   &93.42 & 92.71 & 94.09 & 59.46 &59.83 & 63.01 & 31.18 & 33.28 & 31.74 \\
Fed-Focal Loss    & 94.16 & 92.97 & 94.21 & 53.47 &59.86 & 61.06 & 31.72 & 30.05 & 30.18\\
Ratio Loss   & 93.50 & 92.99 & 94.34 & 59.30 &59.27 & 61.82 & 32.60 & 31.92 &  31.93 \\
CLIMB   & 93.80 & 92.71 & 94.23 & 61.10 & 57.67 & 61.46 & 31.64 & 32.18 & 32.45 \\
CReFF   & 93.94 & 93.85 & 94.76 & 63.25 & 69.36 & 70.30 & 32.20 & 33.46 & 31.60 \\
\midrule
Ours   & \textbf{95.34} & \textbf{95.73} & \textbf{95.93} & \textbf{64.30} & \textbf{71.04} &  \textbf{71.82} &\textbf{32.86} & \textbf{34.63} & \textbf{34.42} \\
\bottomrule
\end{tabular}
\end{center}
\caption{The overall accuracy on the balanced testing sets under different non-i.i.d.\ degrees. Our method consistently outperforms all baselines.}
\label{tab: results of different alpha}
\end{table*}

In our main experiments, we fix the non-i.i.d.\ degree $\alpha=1.0$, in order to mainly explore the effects of different imbalance degrees. Here, we conduct extra experiments with $\alpha=0.1$ and $\alpha=10.0$, and fix the imbalance ratio $\text{IR}=100$. We conduct experiment on MNIST-LT and CIFAR-10 under the full client participation setting, and other experimental settings are kept as the same as that in our main experiments. 

We put the results in Table~\ref{tab: results of different alpha}. We can draw the main conclusion from the table that, \textbf{our method can achieve the best performance under different non-i.i.d.\ degrees}.

\section{Explorations on The Role of The Additional Global Classifier}

As we discussed in Section~\ref{subsec: our optimization target}, in order to address the issue of the contradictory optimization goals on updating $\boldsymbol{W}$ caused by re-balancing the classifier during local training, we add an extra classifier $\widehat{\boldsymbol{W}}$ to help model the global data distribution and make re-balancing $\boldsymbol{W}$ possible. Here, we want to draw comparisons between our proposed two-stream classifier architecture and using one classifier only through experiments on CIFAR-10/100-LT datasets ($\text{IR}=100$, $\alpha=1.0$) under the full client participation situation. The results in the partial client participation setting are similar.

\begin{figure*}[t]
\begin{center}
\subfigure[Results on CIFAR-10-LT.]{ 
\begin{minipage}[t]{0.48\linewidth}  \centerline{\includegraphics[width=1\linewidth]{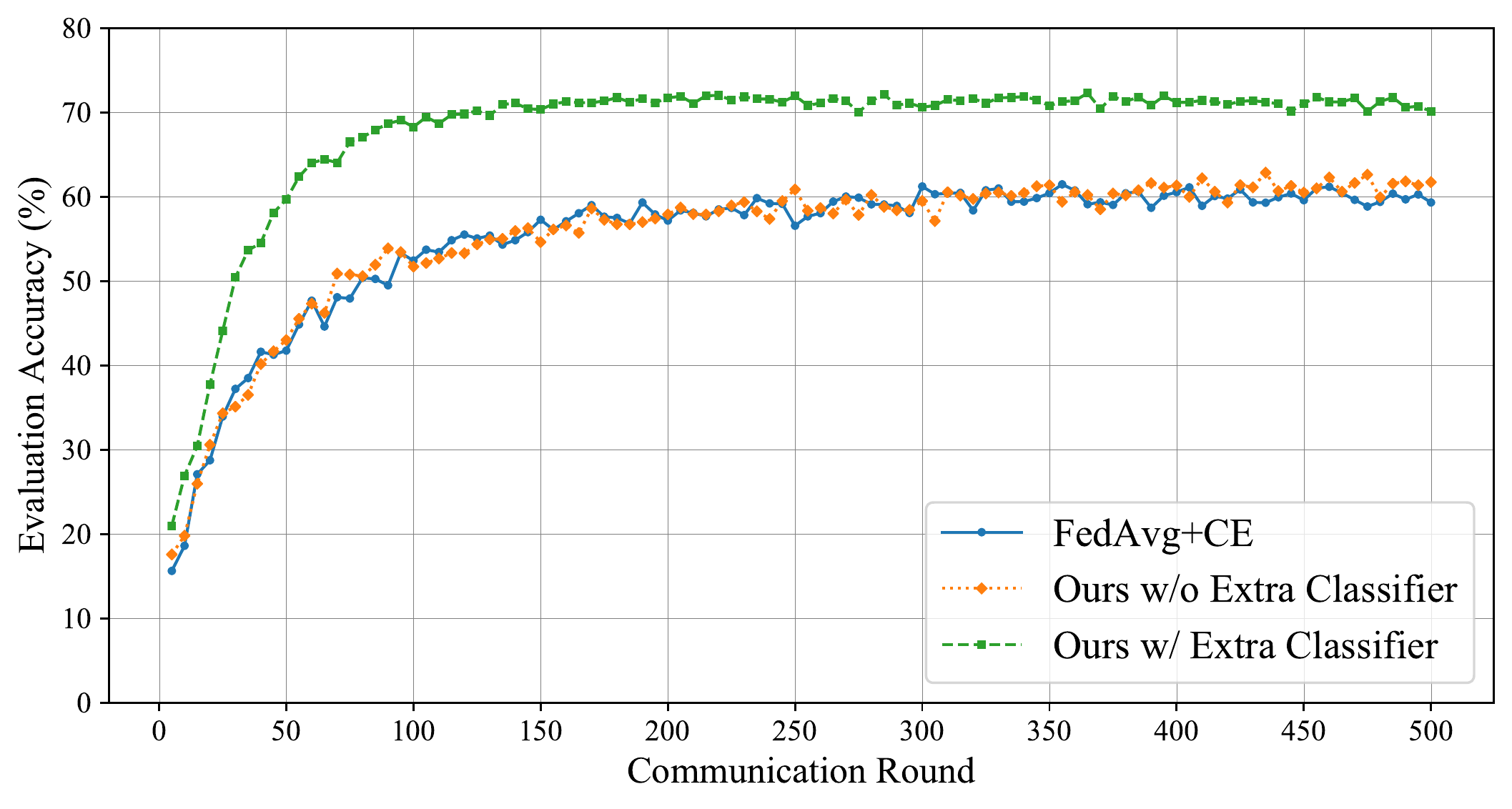}}
\end{minipage}  
}  
\subfigure[Results on CIFAR-100-LT.]{
\begin{minipage}[t]{0.48\linewidth}
\centerline{\includegraphics[width=1\linewidth]{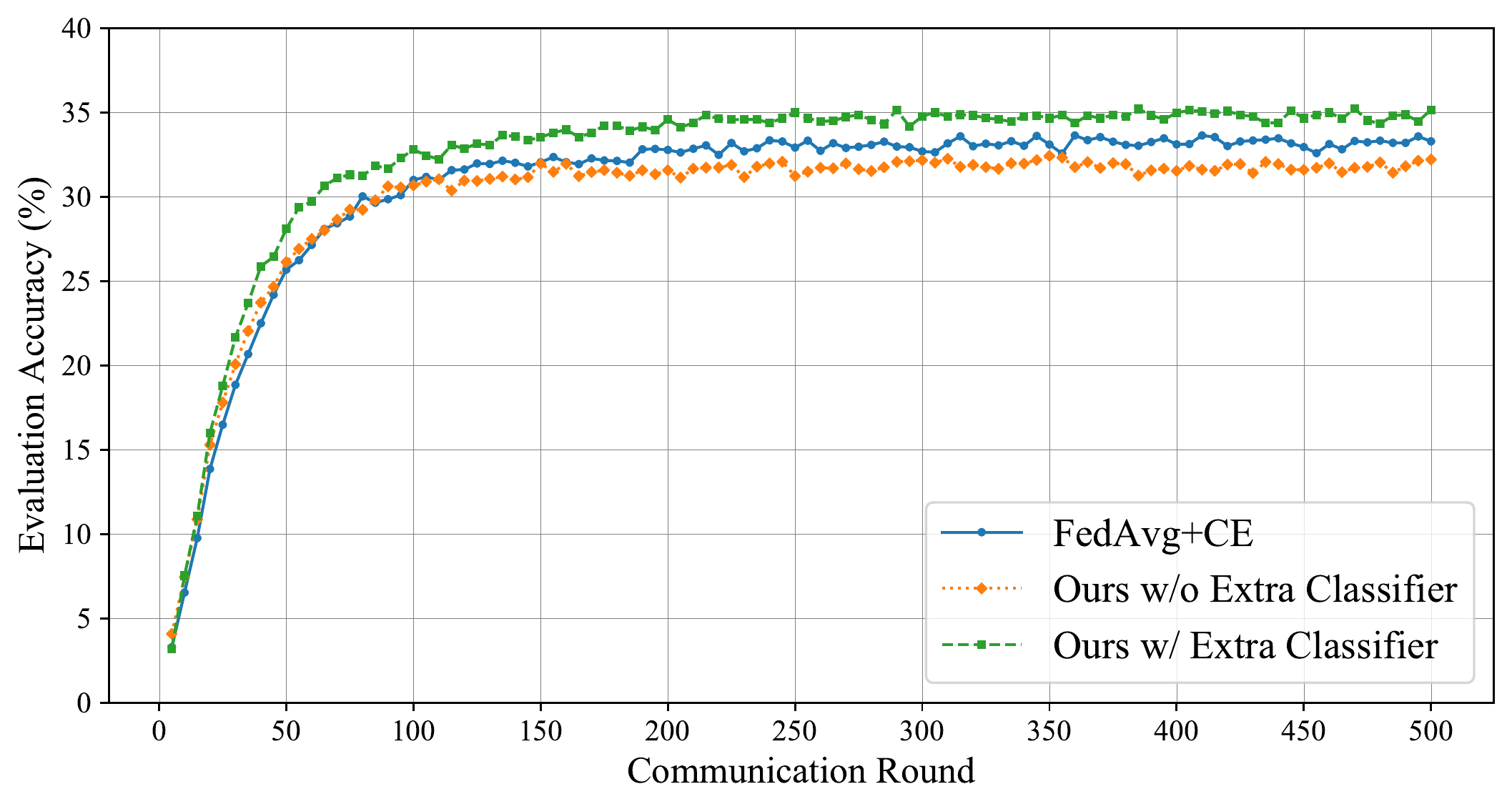}}
\end{minipage}  
}  
\caption{The positive effect of adding an extra classifier in the training phase to help model the global data distribution when locally re-balancing the classifier by our method.}
\label{fig: two-stream classifier}
\end{center}
\end{figure*}

The results are displayed in Figure~\ref{fig: two-stream classifier}. We observe that if we do not add the extra classifier in the training phase, the model will converge to a bad local optimum and behave much worse than that if we adopt the two-stream classifier architecture. \textbf{This helps to validate our motivation and the great effectiveness of introducing a new global classifier to address the optimization difficulty brought by the local classifier re-balancing practice.}

\end{document}